%% file: paper2490.tex
\documentclass[runningheads]{llncs}
\usepackage[bb=boondox]{mathalfa}
\usepackage{xr-hyper}
\usepackage{url}
\usepackage{hyperref}
\usepackage{multirow}
\usepackage{graphicx}
\usepackage{amssymb}
\usepackage{subcaption}
\usepackage{tikz}

\usepackage{amsmath}
\usepackage{bm}
\usepackage{upgreek}
\usepackage{color}

\newcommand{\mf}{\mathbf}

\newcommand{\encoder}{\mathcal{E}_{\bm{\theta}}}
\newcommand{\decoder}{\mathcal{D}_{\bm{\phi}}}

\DeclareMathOperator*{\argmax}{argmax}

\makeatletter
\newcommand\notsotiny{\@setfontsize\notsotiny\@vipt\@viipt}

\newcommand*\samethanks[1][\value{footnote}]{\footnotemark[#1]}

\makeatother

\begin{document}
\title{Intelligent Masking: Deep Q-Learning for Context Encoding in Medical Image Analysis}
\titlerunning{Deep Q-Learning for Context Encoding in Medical Image Analysis}
\author{
Mojtaba Bahrami \inst{1,2,}\thanks{Equal contribution}  \and
Mahsa Ghorbani \inst{1,2,}\samethanks \and
Nassir Navab \inst{1,3}
}

\authorrunning{M. Bahrami et al.}
\institute{Computer Aided Medical Procedures, Technical University of Munich, Germany \and
Sharif University of Technology, Tehran, Iran
 \and
Whiting School of Engineering, Johns Hopkins University, Baltimore, USA\\
}

\maketitle              
\begin{abstract}
The need for a large amount of labeled data in the supervised setting has led recent studies to utilize self-supervised learning to pre-train deep neural networks using unlabeled data. Many self-supervised training strategies have been investigated especially for medical datasets to leverage the information available in the much fewer unlabeled data. One of the fundamental strategies in image-based self-supervision is context prediction. In this approach, a model is trained to reconstruct the contents of an arbitrary missing region of an image based on its surroundings. However, the existing methods adopt a random and blind masking approach by focusing uniformly on all regions of the images. This approach results in a lot of unnecessary network updates that cause the model to forget the rich extracted features. In this work, we develop a novel self-supervised approach that occludes targeted regions to improve the pre-training procedure. To this end, we propose a reinforcement learning-based agent which learns to intelligently mask input images through deep Q-learning. We show that training the agent against the prediction model can significantly improve the semantic features extracted for downstream classification tasks. 
We perform our experiments on two public datasets for diagnosing breast cancer in the ultrasound images and detecting lower-grade glioma with MR images. In our experiments, we show that our novel masking strategy advances the learned features according to the performance on the classification task in terms of accuracy, macro F1, and AUROC.

\end{abstract}
\section{Introduction}
Deep supervised neural networks need a substantial amount of annotated data to demonstrate state-of-the-art performance, particularly in computer vision tasks. However, such an amount of labeled data are hard to acquire especially in medical images. 
Therefore, the methods that do not require large data are becoming more interesting to train high accuracy models with a fewer number of annotated samples. Self-supervision is a family of methods to pre-train deep neural networks using a large number of unlabeled data. The resulting pre-trained networks are then fine-tuned for downstream tasks with much fewer domain-specific labeled samples. Self-supervised strategies showed superior performance not only in learning natural images \cite{doersch2015unsupervised,pathak2016context,noroozi2016unsupervised,gidaris2018unsupervised,li2021cutpaste}, but also for medical images \cite{chen2019self,zhuang2019self,bai2019self,zhang2017self,jamaludin2017self,tajbakhsh2019surrogate,chaitanya2020contrastive} in various tasks including classification, segmentation, anomaly detection, and localization.  \\
In recent years, different self-supervision strategies have been explored that can be classified into two broad categories including \textbf{contrastive learning} and \textbf{pretext task learning} methods. The contrastive learning methods  are based on making the representations of similar images close to each other while making the representations of dissimilar ones further from each other \cite{he2020momentum,chen2020simple,wu2018unsupervised,misra2020self,tian2020contrastive}.
On the other hand, pretext task learning strategies extract auxiliary labels from unlabeled data and then solve a proxy task to learn semantic representations. Pathak et al. \cite{pathak2016context} proposed an in-painting strategy where part of the unlabeled image is masked and a reconstruction model is supposed to predict the masked area. Other methods were also proposed to use the relative positions of image patches \cite{doersch2015unsupervised}, rotation \cite{gidaris2018unsupervised}, and puzzle-solving \cite{noroozi2016unsupervised} as the self-supervision signal. \\
Similar but more customized self-supervised strategies are also explored in the domain of medical image analysis. Jamaludin et al.\cite{jamaludin2017self} proposed the usage of follow-up MR image scans from the same patients in longitudinal studies as a free self-supervision signal. They trained a Siamese CNN to recognize the scans that come from the same patients. In \cite{chen2019self}, the authors proposed a patch swapping strategy followed by a context restoration network to reconstruct the original image. Other methods include pre-training with $2$D sliced image order prediction \cite{zhang2017self} for fine-grained body part recognition, recovering $3$D medical images as a Rubik's cube \cite{zhuang2019self}, and predicting anatomical positions in MR images \cite{bai2019self}. Contrastive learning methods are also investigated in medical imaging. In \cite{tajbakhsh2019surrogate} rotation, reconstruction, and colorization are applied to build similar samples for the contrastive task. A similar work \cite{li2021multi} uses a series of augmentations (Poisson noise, rotation, etc.) for COVID-19 diagnosis. 
However, in the existing methods, the transformations applied in order to build a self-supervision dataset are generally blind with random masking, rotation, swapping, etc. We believe that choosing the appropriate transformations instead of random ones can improve both the efficiency and effectiveness of the training process. This is also more important when dealing with medical images that even contain smaller unlabeled images compared to natural image datasets. In this work, we propose a novel context prediction framework that uses the most informative and effective parts of an image to be masked for our pretext task learning. We developed a Reinforcement Learning (RL) agent that learns the optimal masking strategy against a prediction model that tries to predict the masked region. 
\section{Methods}
The main challenge behind finding the most effective masking strategy is that choosing the mask area is a non-differentiable action that cannot be easily optimized through an end-to-end gradient descent algorithm. The idea developed in this paper is to leverage the capability of reinforcement learning in optimizing the masking strategies to propose intelligent mask areas.
Our method consists of two components, including a Q-learning based masking agent and a prediction model that is pre-trained with our in-painting pretext task. As our masking model is based on deep Q-learning, we first give a brief description of the Q-learning algorithm based on reinforcement learning and then present our main contributions.

\subsection{Deep Q-Learning}
Reinforcement learning is used to train optimal policies where there are no input-output pairs as in the supervised setting or when we need to optimize a policy on a discrete action space. Each environment is modeled by a set of states $S$ and a set of actions $A$ available. At each state $s \in S$ when an agent takes action $a \in A$, then it receives the instant reward $R(s,a)$. A reinforcement setting is modeled by a finite Markov Decision Process (MDP) in which the transition to the next state $s_{t+1}$ only depends on the current state $s_{t}$ and the chosen action $a_{t}$. 
The dynamic of the environment is stated by the transition matrix $T(s, s^\prime, a)$ containing the probability that taking action $a$ at state $s$ will lead to state $s^\prime$ in the next step. The final goal is to find a good policy $\pi(s)$ for the agent that describes which action to pick at each state in order to maximize the total expected return 
$\mathrm{E}\left[\sum_{t=0}^{\infty} \gamma^{t} R(s_t,a_t)\right]$. Here, $\gamma$ is the discount factor that determines the importance of the expected future rewards against the instant rewards.
However, in most real-world problems, $T(s, s^\prime, a)$ is not available. Q-learning \cite{watkins1989learning} is a model-free reinforcement learning algorithm that does not need the dynamics of the environment to be known through a transition matrix. For this purpose a value function $Q(s, a): S \times A \rightarrow \mathbb{R}$ is learned to approximate the expected rewards for an action taken in a given state. To learn the Q-value function, it is first randomly initialized, and then at each step when the agent selects an action, the Q-value function is updated with the following iterative formula that is derived from the Bellman optimality equation:
\begin{equation}
    Q_{new}(s_t, a_t) \leftarrow Q(s_t, a_t)+\alpha \cdot\left[R(s_t,a_t)+\gamma \cdot \max_{a^{\prime}} Q\left(s^{\prime}_{t+1}, a^{\prime}\right)-Q(s_t, a_t)\right]
    \label{eq:q-update}
\end{equation}
where $\alpha$ is the learning rate of the agent.
At each step, the agent corrects its approximation of the total expected future rewards. By updating the Q-function for a sufficient number of iterations the Q-function converges according to \cite{watkins1992q}. Consequently, the optimal policy for the agent can be inferred by taking the action that maximizes the Q-value function at each state.
Furthermore, in deep Q-learning, the Q-value function is approximated by a deep neural network with the state $s_t$ as the input and the Q-values as its output.
\subsection{Self-supervised Intelligent Masking}
\begin{figure}[!tb]
    \centering
    \includegraphics[width=\textwidth,height=0.5\textheight,keepaspectratio]{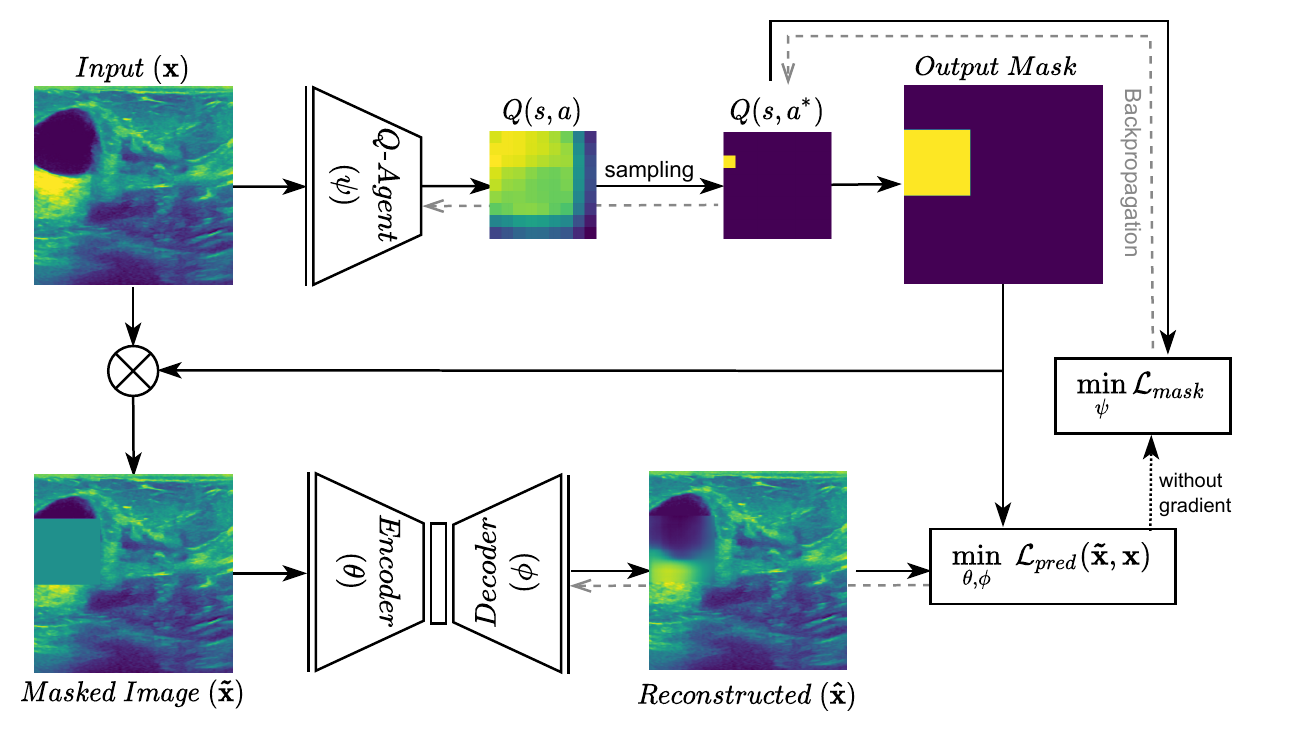}
    \caption{Overview of the proposed model. In the masking network (upper part), the agent selects a patch of the image to be occluded based on its reward estimation of masking each patch. The original image is occluded with the output mask created by the agent. Then, the prediction network (lower part) predicts the occluded region.}
    \label{fig:model}
\end{figure}

Each self-supervision task is composed of two steps: 1) creating paired input-output images and 2) training a prediction model to learn the mapping. In the first step, for a given sample $\mf{x_i}$ from dataset $\mf{X}= \{\mf{x_1}, \mf{x_2}, \dots, \mf{x_n}\}$, we produce an occluded variant $\mf{\tilde{x}_i}$, and then train a prediction model to learn the inverse mapping of $\mf{\tilde{x}_i} \rightarrow \mf{x_i}$. As shown in Fig \ref{fig:model}, we introduce an intelligent masking mechanism to occlude a region based on the input image context to improve the semantic features learned by the prediction model and its performance on subsequent downstream tasks. In the following sections, we present the masking and prediction methods in detail.\\
\textbf{Prediction Network:} According to most self-supervision mechanisms, we also include a prediction model that receives a transformed variant of an image $\mf{\tilde{x}}$ and is pre-trained to reconstruct the original image $\mf{x}$ using the context information. For this purpose, we use a deep convolutional neural network with an Encoder-Decoder architecture to capture the semantics of the input image. The encoder network $\encoder$ consists of a series of stacked Convolution blocks followed by pooling and batch normalization layers with parameters $\bm{\theta}$. The decoder network $\decoder$ with parameters $\bm{\phi}$, includes a set of transposed convolution layers in order to increase the spatial dimension of the latent embedding to generate an output with the same size as the input image. Inspired by \cite{pathak2016context}, we also apply a channel-wise convolutional layer right after the encoder to spread the information through the spatial dimension from one corner to the other in the latent feature map. 
We use pixel-wise Mean Squared Error (MSE) as the loss between the output of the prediction network and the original missing region over the masked area. Therefore the objective of the prediction model is defined as follows:
\begin{align}
    \min_{\bm{\theta,\upvarphi}}  \mathcal{L}_{pred}(\mf{\tilde{x}},\mf{x}) = E_{p(\mf{x})}\|\hat{M} \odot(\mf{x}-\decoder(\encoder(\mf{\tilde{x}})))\|_{2}
\end{align}
where $\odot$ is the element-wise product operation and $\hat{M}$ is the binary output mask.
As stated by \cite{pathak2016context}, although the MSE loss may prefer a blurry reconstruction of the masked region; however, it is sufficient for the feature learning task. \\
\textbf{Masking Network:}
We propose a novel strategy to improve the pretext self-supervision in-painting task by providing intelligent masks instead of random ones in medical images. To this end, we design a mechanism to mask the regions of the input image that the prediction network is weak in its reconstruction. However, because selecting the mask area is a discrete and non-differentiable action, we develop a masking framework based on reinforcement learning to learn the optimal masking policy. To develop such an agent, we design a 1-step episodic problem in which the input image is the current state $s=\mf{x}$ that creates a continuous infinite state space. We also define the set of actions $A$ to be the set of $k \times k$ overlapping image patches that can be selected to be masked, creating a finite action space of size $|A|=k^2$ where $k$ is a hyper-parameter. Finally, we set the loss of the prediction model for the current masked image to be the instant reward for our agent $R(s, a)=\mathcal{L}_{pred}(\mf{\tilde{x}},\mf{x})$. However, unlike the typical reinforcement settings, we only have 1-step long episodes wherein each episode the agent receives the state information (input image $\mf{x}$), takes an action $a^*=\argmax_{a}Q(s=\mf{x}, a)$ to mask an arbitrary region, then the prediction model tries to reconstruct the masked region resulting in the loss value of $\mathcal{L}_{pred}$. After reconstruction, the agent receives the reward $R(s,a)=\mathcal{L}_{pred}$ and the episode terminates. We use Q-learning to train the agent by approximating the Q-value function with a convolutional neural network $Q(s,a,\psi)$ with parameters $\psi$. As the game terminates in the first step, the update rule of Q-values \ref{eq:q-update} becomes as follows:
\begin{equation}
    Q_{new}(s, a) \leftarrow Q(s, a)+\alpha \cdot\left[R(s,a) - Q(s, a)\right]
    \label{eq:q-update}
\end{equation}
where the $R(s, a)$ is set to $\mathcal{L}_{pred}$. In order to train our Q-network $Q(s=\mf{x}, a, \bm{\psi})$ with the above update rule, we define the objective of the masking Deep Q-Network (DQN) according to \cite{hester2018deep} as:
\begin{align}
    \min_{\bm{\psi}}  \mathcal{L}_{mask} = E_{p(\mf{x})} {\|Q(s=\mf{x}, a^*, \bm{\psi}) - \mathcal{L}_{pred}(\mf{\tilde{x}},\mf{x}) \|}^2
\end{align}
where $a^*$ is the action selected to produce the masked variant $\mf{\tilde{x}}$. During our training, to have a balance between exploration and exploitation, we sample the actions based on the distribution of the action scores by applying a soft-max function over the Q-value action scores.



\section{Experiments and Results}
In this section, we evaluate the transferability of our pre-trained network to the disease classification task, using the encoder $\encoder$ as a feature extractor. We experiment on two datasets: Low-Grade Gliomas (LGGs) \cite{buda2019association,mazurowski2017radiogenomics} and Breast UltraSound Images (BUSI)  dataset\cite{al2020dataset} in section \ref{subsec:lgg} and section \ref{subsec:busi}, respectively. In section \ref{subsec:vis}, we provide a qualitative example comparing other feature learning strategies.
We compare our method with one unsupervised and two other self-supervised strategies. They include an unsupervised auto-encoder (\textit{Reconstruction}) \cite{bengio2006greedy}, context prediction (in-painting) with random masking (\textit{Context Prediction}) \cite{pathak2016context}, and self-supervision using disordering random patches and restoring the original image (\textit{Context Restoration}) \cite{chen2019self}. 
For both datasets, the whole training set is used for the self-supervision task. The performance of the trained classifier is evaluated in terms of accuracy, macro F1 score, and AUROC. We compare the performance of the pre-training phase in feature extraction especially when the training samples are limited. \\
\textbf{Implementation Details:} We start the training of both the self-supervised model and the classifier with the learning rate of $10^{-4}$ and use a multi-step scheduler that decays the learning rate by a factor of $0.3$ in two equal milestones during both self-supervision and classification training. We utilize dropout with the rate of $p=0.3$ just for fully-connected layers in the classifier and Adam optimizer \cite{da2014method} is selected for the training of the networks. More details of the implementation can be found in the supplementary materials.
\subsection{Diagnosis of Lower-Grade Glioma}
\label{subsec:lgg}
In our first experiment, we model the presence of the abnormality in the brain MR images as a binary classification with the purpose of Lower-Grade Glioma (LGG) detection. To this aim, we use Lower-Grade Glioma (LGG) dataset \cite{buda2019association,mazurowski2017radiogenomics}  obtained from The Cancer Imaging Archive (TCIA). The dataset contains $3929$ MR images corresponding to $110$ patients where $2556$ images are normal and $1373$ images are with abnormalities. We take $10\%$ of the data for validation and $10\%$ for testing. \\
\textbf{Results and Analysis:} Table \ref{tab:res-lgg} shows the average accuracy, macro F1, and AUROC over 5 runs with different random initializations. We evaluate the classification performance of all methods for three different training set sizes containing 961, 480, and 192 samples in each. Our method outperforms the rest in all metrics and all training set sizes with a higher margin for lower training set sizes showing the robustness of the pre-trained model. Other methods have similar performances in large training sets; however, they perform differently for lower training sizes. From competitor methods, Context Prediction obtains better results than others.

\begin{table}[!htb]
\centering
\caption{Results of the compared methods on LGG dataset}
\footnotesize
\begin{tabular}{|c|c|c|c|c|}
\hline
Training data & Method & Accuracy & Macro F1 & AUROC \\ \hline \hline 
\multirow{4}{*}{192}  & Reconstruction & $ 0.682 \pm 0.0218 $ & $ 0.676 \pm 0.0237 $ & $ 0.763 \pm 0.0362 $ \\  
 & Context Restoration & $ 0.69 \pm 0.0226 $ & $ 0.686 \pm 0.0237 $ & $ 0.768 \pm 0.037 $ \\  
 & Context Prediction & $ 0.732 \pm 0.0063 $ & $ 0.731 \pm 0.0063 $ & $ 0.81 \pm 0.0179 $ \\  
 & Intelligent-Masking & $\mathbf{ 0.764 \pm 0.0168} $ & $\mathbf{ 0.761 \pm 0.0176} $ & $\mathbf{ 0.819 \pm 0.0103} $ \\  
 \hline 
\multirow{4}{*}{480}  & Reconstruction & $ 0.744 \pm 0.0241 $ & $ 0.743 \pm 0.0246 $ & $ 0.812 \pm 0.0322 $ \\  
 & Context Restoration & $ 0.75 \pm 0.0059 $ & $ 0.749 \pm 0.006 $ & $ 0.818 \pm 0.0279 $ \\  
 & Context Prediction & $ 0.754 \pm 0.0113 $ & $ 0.753 \pm 0.0109 $ & $ 0.832 \pm 0.0205 $ \\  
 & Intelligent-Masking & $\mathbf{ 0.782 \pm 0.0075} $ & $ \mathbf{0.78 \pm 0.0073 }$ & $\mathbf{ 0.855 \pm 0.0165} $ \\  
 \hline 
\multirow{4}{*}{961}  & Reconstruction & $ 0.764 \pm 0.0254 $ & $ 0.764 \pm 0.0256 $ & $ 0.822 \pm 0.0291 $ \\  
 & Context Restoration & $ 0.768 \pm 0.0145 $ & $ 0.767 \pm 0.0146 $ & $ 0.826 \pm 0.0269 $ \\  
 & Context Prediction & $ 0.762 \pm 0.0104 $ & $ 0.762 \pm 0.0099 $ & $ 0.842 \pm 0.0208 $ \\  
 & Intelligent-Masking & $\mathbf{0.775 \pm 0.013} $ & $\mathbf{ 0.774 \pm 0.0126} $ & $ \mathbf{0.855 \pm 0.0197} $ \\  
 \hline 
\end{tabular}
\label{tab:res-lgg}
\end{table}

\subsection{Breast Cancer Detection}
\label{subsec:busi}
Detection of breast cancer and its type is our second experiment. Breast UltraSound Images (BUSI) dataset \cite{al2020dataset} includes $2$D ultrasound images collected from women between ages $25$ and $75$ years old in $2018$. This dataset provides $780$ single-channel grayscale ultrasound images categorized into $133$ normal, and $647$ cancer cases.
We split the data into $50\%$, $25\%$, and $25\%$ for training, validation, and test, respectively. \\
\textbf{Results and Analysis:} Table \ref{tab:res-busi} shows the average accuracy, macro F1, and AUROC over 5 runs with different random initializations. We compare the performance of the methods on three different training sizes. However, unlike the dataset in section \ref{subsec:lgg}, there are much fewer samples in each setup in this dataset. The results show that Intelligent-Masking is much more superior in relatively smaller training annotations. We observe that Context Restoration shows lower results compared to Context Prediction which might be due to the fact that our classification tasks need the pre-trained network focus on the local structures like tumor regions rather than the global information. However, the loss of the Context Restoration is defined globally over the entire image unlike the Context Prediction and Intelligent-Masking that define the loss only over the masked regions.

\begin{table}[!htb]
\centering
\caption{Results of the compared methods on BUSI dataset}
\begin{tabular}{|c|c|c|c|c|}
\hline
Training data & Method & Accuracy & Macro F1 & AUROC \\ \hline \hline 
\multirow{4}{*}{13}  & Reconstruction & $ 0.576 \pm 0.063 $ & $ 0.521 \pm 0.1089 $ & $ 0.616 \pm 0.0179 $ \\  
 & Context Restoration & $ 0.675 \pm 0.0067 $ & $ 0.658 \pm 0.008 $ & $ 0.668 \pm 0.0127 $ \\  
 & Context Prediction & $ 0.752 \pm 0.0344 $ & $ 0.744 \pm 0.0425 $ & $ 0.801 \pm 0.003 $ \\  
 & Intelligent-Masking & $\mathbf{0.776 \pm 0.0106} $ & $\mathbf{0.775 \pm 0.0103}$ & $\mathbf{ 0.857 \pm 0.0079 }$ \\  
 \hline 
\multirow{4}{*}{66}  & Reconstruction & $ 0.585 \pm 0.0645 $ & $ 0.579 \pm 0.0675 $ & $ 0.667 \pm 0.0973 $ \\  
 & Context Restoration & $ 0.716 \pm 0.0183 $ & $ 0.709 \pm 0.0187 $ & $ 0.773 \pm 0.0095 $ \\  
 & Context Prediction & $ 0.779 \pm 0.0221 $ & $ 0.777 \pm 0.0213 $ & $ 0.828 \pm 0.0116 $ \\  
 & Intelligent-Masking & $ \mathbf{0.794 \pm 0.0221} $ & $ \mathbf{0.79 \pm 0.024} $ & $ \mathbf{0.862 \pm 0.0058} $ \\  
 \hline 
\multirow{4}{*}{129}  & Reconstruction & $ 0.639 \pm 0.0267 $ & $ 0.638 \pm 0.0272 $ & $ 0.702 \pm 0.0098 $ \\  
 & Context Restoration & $ 0.71 \pm 0.0403 $ & $ 0.707 \pm 0.0366 $ & $ 0.788 \pm 0.0067 $ \\  
 & Context Prediction & $ 0.776 \pm 0.0236 $ & $ 0.774 \pm 0.0249 $ & $ 0.827 \pm 0.0062 $ \\  
 & Intelligent-Masking & $ \mathbf{0.8 \pm 0.0309} $ & $ \mathbf{0.798 \pm 0.0332} $ & $ \mathbf{0.865 \pm 0.005} $ \\  
 \hline 
\end{tabular}
\label{tab:res-busi}
\end{table}

\subsection{Qualitative Analysis}
\label{subsec:vis}
\begin{figure*}[!htb]
    \begin{subfigure}[t]{0.24\textwidth} 
        \includegraphics[width=\textwidth]{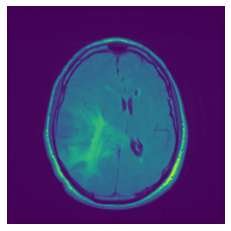}
    \end{subfigure}
 \begin{tikzpicture}
        \tikz{\draw[thin,black, dashed](0,0.05) -- (0,3);} 
    \end{tikzpicture}
    \begin{subfigure}[t]{0.24\textwidth}
        \includegraphics[width=\textwidth]{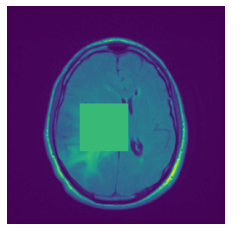}
    \end{subfigure}
    \begin{subfigure}[t]{0.24\textwidth}
        \includegraphics[width=\textwidth]{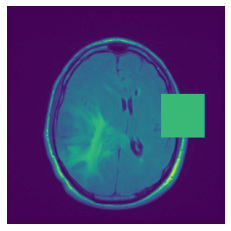}
    \end{subfigure}
    \begin{subfigure}[t]{0.24\textwidth}
        \includegraphics[width=\textwidth]{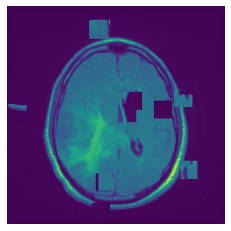}
    \end{subfigure} \\
    \begin{subfigure}[t]{0.24\textwidth}
        \includegraphics[width=\textwidth]{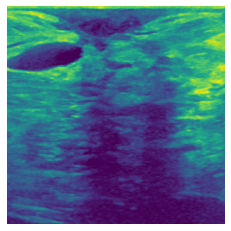}
        \caption{Original}
    \end{subfigure}
 \begin{tikzpicture}
        \tikz{\draw[thin,black, dashed](0,0.05) -- (0,3);} 
    \end{tikzpicture}
    \begin{subfigure}[t]{0.24\textwidth}
        \includegraphics[width=\textwidth]{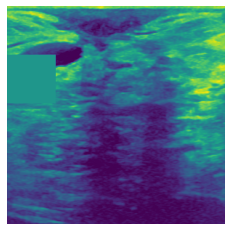}
        \caption{Intelligent-Masking}
    \end{subfigure}
    \begin{subfigure}[t]{0.24\textwidth}
        \includegraphics[width=\textwidth]{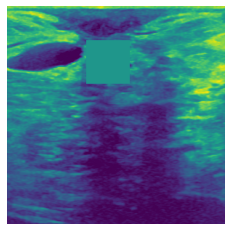}
        \caption{Context Prediction}
    \end{subfigure}
    \begin{subfigure}[t]{0.24\textwidth}
        \includegraphics[width=\textwidth]{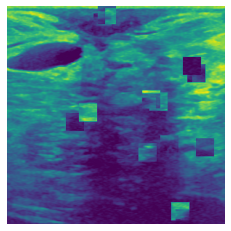}
        \caption{Context Restoration}
    \end{subfigure} 
  \caption{Qualitative examples of compared method's strategies for masking}
  \label{fig:res-vis}

\end{figure*}
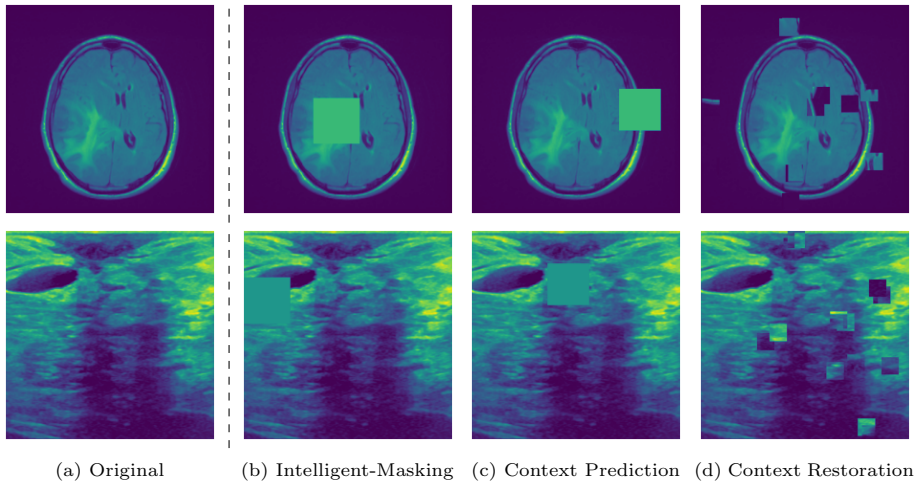

In figure \ref{fig:res-vis}, we present qualitative examples of different masking strategies. It is observed that, unlike context prediction and restoration, our method tends to propose targeted masks like the tumor regions or regions with abnormalities and avoids masking less helpful regions. However, it should be noted that Intelligent-Masking does not necessarily mask the tumor regions but considers all areas of interest that results in better feature learning. Examples of other masking samples are provided in supplementary materials. Furthermore, in medical images, unlike natural scenes, the structures are very local with imbalanced information throughout an image. Therefore, random masking strategies as shown in Fig \ref{fig:res-vis} operate ineffectively by masking non-informative regions.

\section{Discussion $\&$ Conclusion}
This paper proposes an intelligent masking strategy to train rich feature representation instead of random masking. To this aim, we leverage a reinforcement learning-based agent which learns to intelligently mask input images through deep Q-learning. To evaluate our model, we use the trained encoder for the classification task on two medical image datasets for the diagnosis of lower-grade glioma and breast cancer. Experimental results show that Intelligent-Masking has a consistent enhancement in terms of accuracy, macro F1, and AUROC compared to the SOTA methods. We believe that this work opens a path for enhancing self-supervision strategies like rotation and patch swapping and making them more context-aware. 

%
%
\bibliographystyle{splncs04}
\bibliography{ref}
\newpage
\section{Supplementary Material}
\input{paper2490-supplement_raw}
\end{document}

%% file: paper2490-supplement_raw.tex
\textbf{Implementation Details:} In our first experiment (detecting lower-grade glioma with MR images), we run 4 different convolutional architectures for each method and find the best architecture for each one based on macro F1 of the validation set. These  architectures include:
\\
\begin{enumerate}
\item Auto-Encoder: 
    \begin{itemize}
        \item[$\bullet$] Encoder channels:(16, 32, 64, 64, 128, 128)
        \item[$\bullet$] Decoder channels: (128, 128, 64, 64, 32, 16)
    \end{itemize}
\item Auto-Encoder: 
    \begin{itemize}
        \item[$\bullet$] Encoder channels:(32, 64, 64, 128, 128, 256)
        \item[$\bullet$] Decoder channels: (256, 128, 64, 64, 64, 32)
    \end{itemize}
\item UNet: 
    \begin{itemize}
        \item[$\bullet$] Encoder channels:(16, 32, 64, 64, 128, 128)
        \item[$\bullet$] Decoder channels: (128, 128, 64, 64, 32, 16)
    \end{itemize}
\item UNet: 
    \begin{itemize}
        \item[$\bullet$] Encoder channels:(32, 64, 64, 128, 128, 256)
        \item[$\bullet$] Decoder channels: (256, 128, 64, 64, 64, 32)
    \end{itemize}
\end{enumerate}

In the second experiment (diagnosing breast cancer in the ultrasound images), unlike the previous experiment we use one equal architecture for all the comparing methods to show our superiority with the same model type and number of parameters. The architecture used is:

\begin{itemize}
    \item[$-$] Auto-Encoder
    \begin{itemize}
        \item[$\bullet$] Encoder channels:(16, 32, 64, 64, 128, 128)
        \item[$\bullet$] Decoder channels: (128, 128, 64, 64, 32, 16)
    \end{itemize}
\end{itemize}

In both experiments we use a Q-Agent network with the same network size of the encoder.

\begin{figure*}[!htb]
    \begin{subfigure}[t]{0.24\textwidth} 
        \includegraphics[width=\textwidth]{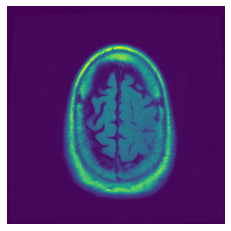}
    \end{subfigure}
 \begin{tikzpicture}
        \tikz{\draw[thin,black, dashed](0,0.05) -- (0,3);} 
    \end{tikzpicture}
    \begin{subfigure}[t]{0.24\textwidth}
        \includegraphics[width=\textwidth]{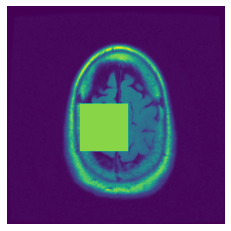}
    \end{subfigure}
    \begin{subfigure}[t]{0.24\textwidth}
        \includegraphics[width=\textwidth]{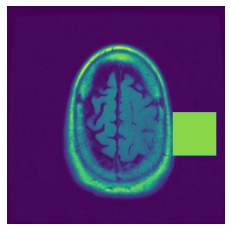}
    \end{subfigure}
    \begin{subfigure}[t]{0.24\textwidth}
        \includegraphics[width=\textwidth]{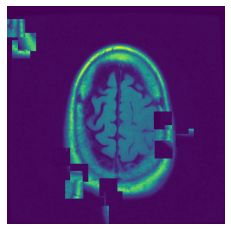}
    \end{subfigure} \\
    \begin{subfigure}[t]{0.24\textwidth}
        \includegraphics[width=\textwidth]{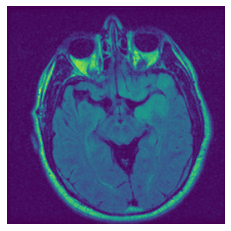}
    \end{subfigure}
     \begin{tikzpicture}
        \tikz{\draw[thin,black, dashed](0,0.05) -- (0,3);} 
    \end{tikzpicture}
    \begin{subfigure}[t]{0.24\textwidth}
        \includegraphics[width=\textwidth]{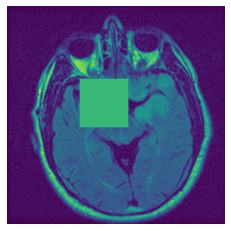}
    \end{subfigure}
    \begin{subfigure}[t]{0.24\textwidth}
        \includegraphics[width=\textwidth]{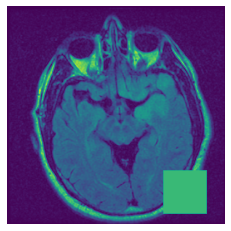}
    \end{subfigure}
    \begin{subfigure}[t]{0.24\textwidth}
        \includegraphics[width=\textwidth]{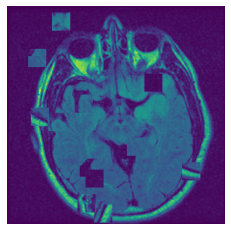}
    \end{subfigure} \\
    \begin{subfigure}[t]{0.24\textwidth}
        \includegraphics[width=\textwidth]{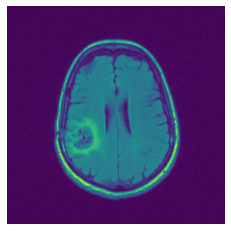}
    \end{subfigure}
 \begin{tikzpicture}
        \tikz{\draw[thin,black, dashed](0,0.05) -- (0,3);} 
    \end{tikzpicture}
    \begin{subfigure}[t]{0.24\textwidth}
        \includegraphics[width=\textwidth]{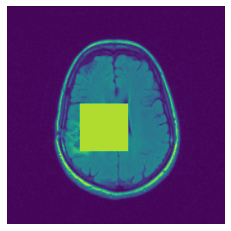}
    \end{subfigure}
    \begin{subfigure}[t]{0.24\textwidth}
        \includegraphics[width=\textwidth]{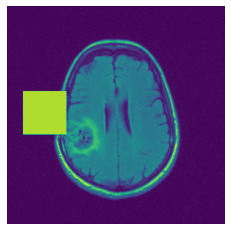}
    \end{subfigure}
    \begin{subfigure}[t]{0.24\textwidth}
        \includegraphics[width=\textwidth]{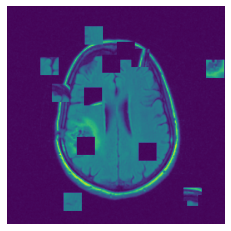}
    \end{subfigure} 

    \begin{subfigure}[t]{0.24\textwidth} 
        \includegraphics[width=\textwidth]{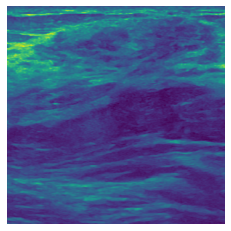}
    \end{subfigure}
 \begin{tikzpicture}
        \tikz{\draw[thin,black, dashed](0,0.05) -- (0,3);} 
    \end{tikzpicture}
    \begin{subfigure}[t]{0.24\textwidth}
        \includegraphics[width=\textwidth]{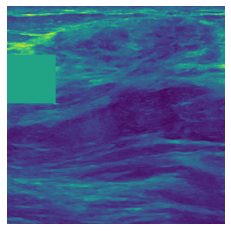}
    \end{subfigure}
    \begin{subfigure}[t]{0.24\textwidth}
        \includegraphics[width=\textwidth]{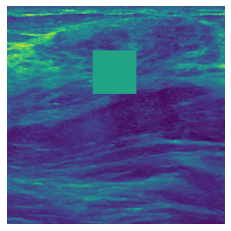}
    \end{subfigure}
    \begin{subfigure}[t]{0.24\textwidth}
        \includegraphics[width=\textwidth]{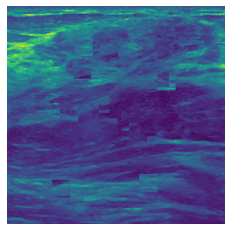}
    \end{subfigure} \\
    \begin{subfigure}[t]{0.24\textwidth}
        \includegraphics[width=\textwidth]{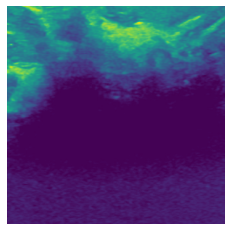}
    \end{subfigure}
     \begin{tikzpicture}
        \tikz{\draw[thin,black, dashed](0,0.05) -- (0,3);} 
    \end{tikzpicture}
    \begin{subfigure}[t]{0.24\textwidth}
        \includegraphics[width=\textwidth]{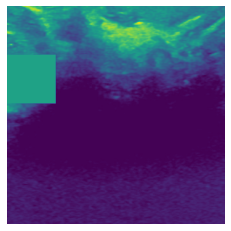}
    \end{subfigure}
    \begin{subfigure}[t]{0.24\textwidth}
        \includegraphics[width=\textwidth]{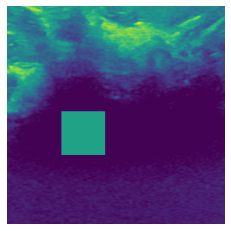}
    \end{subfigure}
    \begin{subfigure}[t]{0.24\textwidth}
        \includegraphics[width=\textwidth]{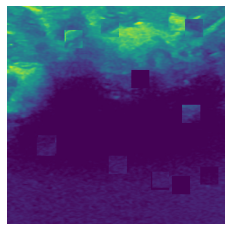}
    \end{subfigure} \\
    \begin{subfigure}[t]{0.24\textwidth}
        \includegraphics[width=\textwidth]{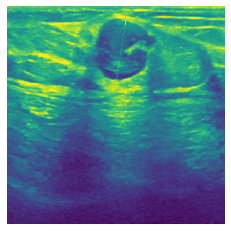}
        \caption{Original}
    \end{subfigure}
 \begin{tikzpicture}
        \tikz{\draw[thin,black, dashed](0,0.05) -- (0,3);} 
    \end{tikzpicture}
    \begin{subfigure}[t]{0.24\textwidth}
        \includegraphics[width=\textwidth]{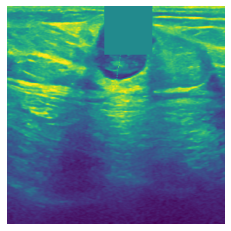}
        \caption{Intelligent-Masking}
    \end{subfigure}
    \begin{subfigure}[t]{0.24\textwidth}
        \includegraphics[width=\textwidth]{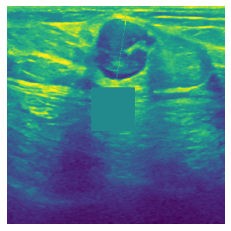}
        \caption{Context Prediction}
    \end{subfigure}
    \begin{subfigure}[t]{0.24\textwidth}
        \includegraphics[width=\textwidth]{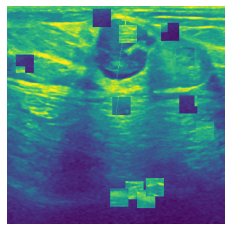}
        \caption{Context Restoration}
    \end{subfigure} 
  \caption{More qualitative examples of different distorting strategies including our method. We show the different self-supervised mechanisms on both datasets of MR (rows: 1-3) and ultrasound (rows: 4-6) images for lower-grade glioma and breast cancer diagnosis respectively. We include both images of normal (rows: 1,4) and cancer (rows: 2,3,5,6) conditions for each dataset. We also observe that our method treats each image based on its context information with no predetermined strategy.}
  \label{fig:res-vis}

\end{figure*}
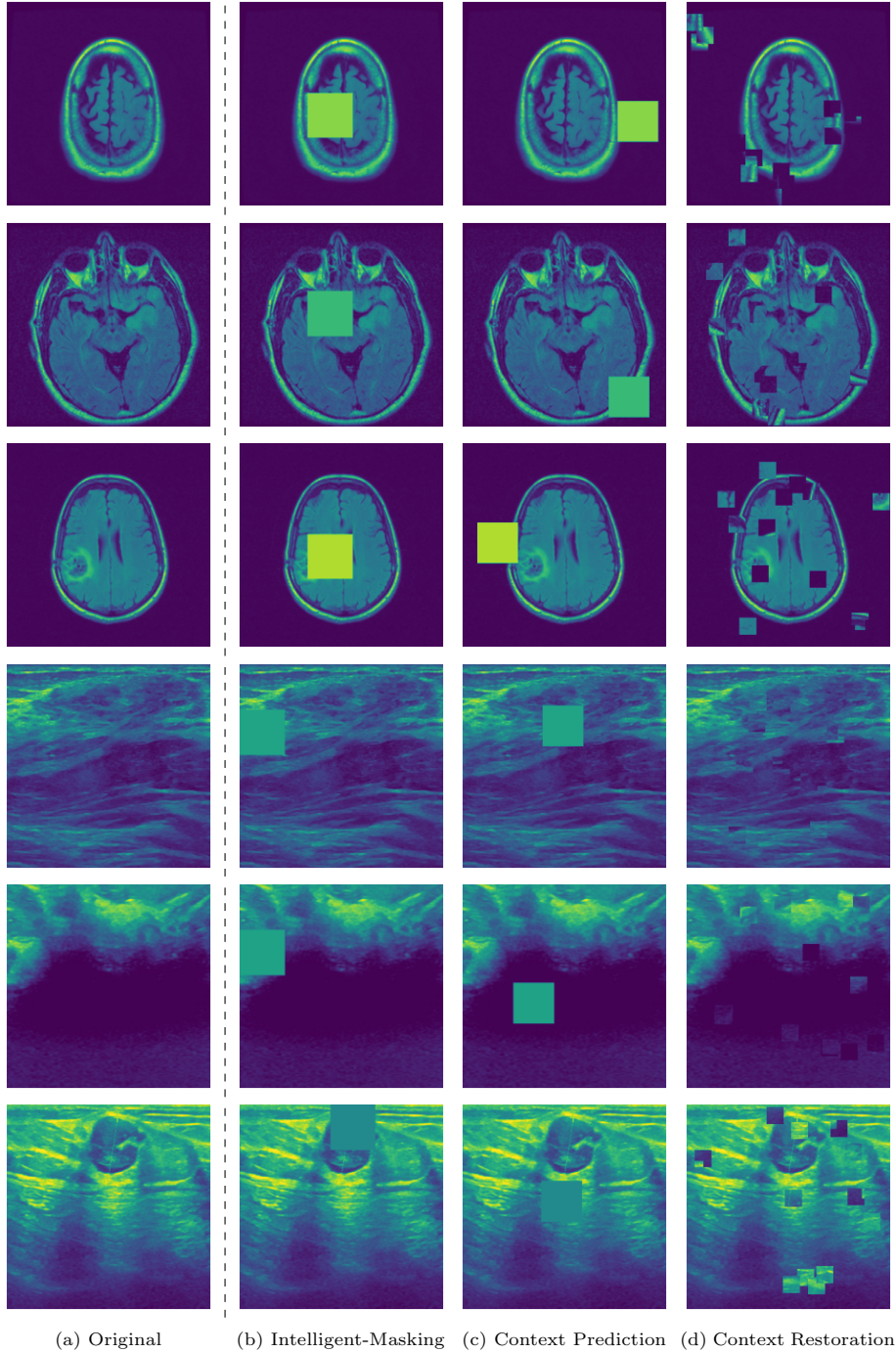

%% file: paper2490.bbl
\begin{thebibliography}{10}
\providecommand{\url}[1]{\texttt{#1}}
\providecommand{\urlprefix}{URL }
\providecommand{\doi}[1]{https://doi.org/#1}

\bibitem{al2020dataset}
Al-Dhabyani, W., Gomaa, M., Khaled, H., Fahmy, A.: Dataset of breast ultrasound
  images. Data in brief  \textbf{28},  104863 (2020)

\bibitem{bai2019self}
Bai, W., Chen, C., Tarroni, G., Duan, J., Guitton, F., Petersen, S.E., Guo, Y.,
  Matthews, P.M., Rueckert, D.: Self-supervised learning for cardiac mr image
  segmentation by anatomical position prediction. In: International Conference
  on Medical Image Computing and Computer-Assisted Intervention. pp. 541--549.
  Springer (2019)

\bibitem{bengio2006greedy}
Bengio, Y., Lamblin, P., Popovici, D., Larochelle, H.: Greedy layer-wise
  training of deep networks. Advances in neural information processing systems
  \textbf{19} (2006)

\bibitem{buda2019association}
Buda, M., Saha, A., Mazurowski, M.A.: Association of genomic subtypes of
  lower-grade gliomas with shape features automatically extracted by a deep
  learning algorithm. Computers in biology and medicine  \textbf{109},
  218--225 (2019)

\bibitem{chaitanya2020contrastive}
Chaitanya, K., Erdil, E., Karani, N., Konukoglu, E.: Contrastive learning of
  global and local features for medical image segmentation with limited
  annotations. Advances in Neural Information Processing Systems  \textbf{33},
  12546--12558 (2020)

\bibitem{chen2019self}
Chen, L., Bentley, P., Mori, K., Misawa, K., Fujiwara, M., Rueckert, D.:
  Self-supervised learning for medical image analysis using image context
  restoration. Medical image analysis  \textbf{58},  101539 (2019)

\bibitem{chen2020simple}
Chen, T., Kornblith, S., Norouzi, M., Hinton, G.: A simple framework for
  contrastive learning of visual representations. In: International conference
  on machine learning. pp. 1597--1607. PMLR (2020)

\bibitem{da2014method}
Da, K.: A method for stochastic optimization. arXiv preprint arXiv:1412.6980
  (2014)

\bibitem{doersch2015unsupervised}
Doersch, C., Gupta, A., Efros, A.A.: Unsupervised visual representation
  learning by context prediction. In: Proceedings of the IEEE international
  conference on computer vision. pp. 1422--1430 (2015)

\bibitem{gidaris2018unsupervised}
Gidaris, S., Singh, P., Komodakis, N.: Unsupervised representation learning by
  predicting image rotations. arXiv preprint arXiv:1803.07728  (2018)

\bibitem{he2020momentum}
He, K., Fan, H., Wu, Y., Xie, S., Girshick, R.: Momentum contrast for
  unsupervised visual representation learning. In: Proceedings of the IEEE/CVF
  conference on computer vision and pattern recognition. pp. 9729--9738 (2020)

\bibitem{hester2018deep}
Hester, T., Vecerik, M., Pietquin, O., Lanctot, M., Schaul, T., Piot, B.,
  Horgan, D., Quan, J., Sendonaris, A., Osband, I., et~al.: Deep q-learning
  from demonstrations. In: Proceedings of the AAAI Conference on Artificial
  Intelligence. vol.~32 (2018)

\bibitem{jamaludin2017self}
Jamaludin, A., Kadir, T., Zisserman, A.: Self-supervised learning for spinal
  mris. In: Deep Learning in Medical Image Analysis and Multimodal Learning for
  Clinical Decision Support, pp. 294--302. Springer (2017)

\bibitem{li2021cutpaste}
Li, C.L., Sohn, K., Yoon, J., Pfister, T.: Cutpaste: Self-supervised learning
  for anomaly detection and localization. In: Proceedings of the IEEE/CVF
  Conference on Computer Vision and Pattern Recognition. pp. 9664--9674 (2021)

\bibitem{li2021multi}
Li, J., Zhao, G., Tao, Y., Zhai, P., Chen, H., He, H., Cai, T.: Multi-task
  contrastive learning for automatic ct and x-ray diagnosis of covid-19.
  Pattern Recognition  \textbf{114},  107848 (2021)

\bibitem{mazurowski2017radiogenomics}
Mazurowski, M.A., Clark, K., Czarnek, N.M., Shamsesfandabadi, P., Peters, K.B.,
  Saha, A.: Radiogenomics of lower-grade glioma: algorithmically-assessed tumor
  shape is associated with tumor genomic subtypes and patient outcomes in a
  multi-institutional study with the cancer genome atlas data. Journal of
  neuro-oncology  \textbf{133}(1),  27--35 (2017)

\bibitem{misra2020self}
Misra, I., Maaten, L.v.d.: Self-supervised learning of pretext-invariant
  representations. In: Proceedings of the IEEE/CVF Conference on Computer
  Vision and Pattern Recognition. pp. 6707--6717 (2020)

\bibitem{noroozi2016unsupervised}
Noroozi, M., Favaro, P.: Unsupervised learning of visual representations by
  solving jigsaw puzzles. In: European conference on computer vision. pp.
  69--84. Springer (2016)

\bibitem{pathak2016context}
Pathak, D., Krahenbuhl, P., Donahue, J., Darrell, T., Efros, A.A.: Context
  encoders: Feature learning by inpainting. In: Proceedings of the IEEE
  conference on computer vision and pattern recognition. pp. 2536--2544 (2016)

\bibitem{tajbakhsh2019surrogate}
Tajbakhsh, N., Hu, Y., Cao, J., Yan, X., Xiao, Y., Lu, Y., Liang, J.,
  Terzopoulos, D., Ding, X.: Surrogate supervision for medical image analysis:
  Effective deep learning from limited quantities of labeled data. In: 2019
  IEEE 16th International Symposium on Biomedical Imaging (ISBI 2019). pp.
  1251--1255. IEEE (2019)

\bibitem{tian2020contrastive}
Tian, Y., Krishnan, D., Isola, P.: Contrastive multiview coding. In: European
  conference on computer vision. pp. 776--794. Springer (2020)

\bibitem{watkins1992q}
Watkins, C.J., Dayan, P.: Q-learning. Machine learning  \textbf{8}(3),
  279--292 (1992)

\bibitem{watkins1989learning}
Watkins, C.J.C.H.: Learning from delayed rewards  (1989)

\bibitem{wu2018unsupervised}
Wu, Z., Xiong, Y., Yu, S.X., Lin, D.: Unsupervised feature learning via
  non-parametric instance discrimination. In: Proceedings of the IEEE
  conference on computer vision and pattern recognition. pp. 3733--3742 (2018)

\bibitem{zhang2017self}
Zhang, P., Wang, F., Zheng, Y.: Self supervised deep representation learning
  for fine-grained body part recognition. In: 2017 IEEE 14th International
  Symposium on Biomedical Imaging (ISBI 2017). pp. 578--582. IEEE (2017)

\bibitem{zhuang2019self}
Zhuang, X., Li, Y., Hu, Y., Ma, K., Yang, Y., Zheng, Y.: Self-supervised
  feature learning for 3d medical images by playing a rubik’s cube. In:
  International Conference on Medical Image Computing and Computer-Assisted
  Intervention. pp. 420--428. Springer (2019)

\end{thebibliography}
